# Benchmarking Retrieval Strategies for Biomedical Retrieval-Augmented Generation: A Controlled Empirical Study[1]

**Devi Prasad Bal**
The University of Texas at Austin
devibal@utexas.edu

**Subhashree Puhan**
The University of Texas at Austin
subhashree.puhan@utexas.edu

## Abstract

Retrieval-Augmented Generation (RAG) offers a well-established path to grounding large language model (LLM) outputs in external knowledge, yet the question of which retrieval strategy works best in a high-stakes domain such as biomedicine has not received the controlled, multi-metric treatment it deserves. This paper presents a systematic empirical comparison of five retrieval strategies—Dense Vector Search, Hybrid BM25 + Dense retrieval, Cross-Encoder Reranking, Multi-Query Expansion, and Maximal Marginal Relevance (MMR)—within a biomedical question-answering RAG pipeline. All strategies share a fixed generation model (GPT-4o-mini), a common vector store (ChromaDB), and OpenAI's text-embedding-3-small embeddings, ensuring that observed differences are attributable to retrieval alone. Evaluation is conducted on 250 question-answer (QA) pairs drawn from a preprocessed subset of the BioASQ benchmark (rag-mini-bioasq) using four DeepEval metrics: contextual precision, contextual recall, faithfulness, and answer relevancy, each reported with 95 % confidence intervals computed over 250 samples. A no-context ablation is included as a lower bound. Cross-Encoder Reranking achieves the best composite score (0.827) and highest contextual precision (0.852), confirming that query-document interaction yields measurable retrieval gains. Multi-Query Expansion, despite its recall-oriented design, produces the weakest contextual precision (0.671), suggesting naive query diversification introduces retrieval noise. MMR sacrifices answer relevancy for diversity, while the Dense baseline (composite 0.822) falls within 0.005 points of the top strategy. All RAG conditions dramatically outperform the no-context ablation on answer relevancy (0.658–0.701 vs. 0.287), confirming the practical value of retrieval. The full pipeline, hyperparameters, and evaluation code are publicly available.

## 1 Introduction

Large language models have achieved remarkable performance across a wide range of natural language tasks, yet two fundamental weaknesses remain stubbornly persistent: their knowledge is frozen at training time, and they are prone to generating plausible-sounding but factually incorrect statements—a phenomenon commonly called hallucination [1, 2, 3]. In low-stakes settings these limitations are inconvenient; in biomedical and clinical contexts they are potentially dangerous. When a system is asked about drug interactions, diagnostic criteria, or treatment protocols, accuracy is not a design aspiration—it is a patient-safety requirement.

[1] Code: https://github.com/deviprasadbal/RAGHealthcareRetrievalStrategies

Retrieval-Augmented Generation (RAG) addresses both problems by coupling the LLM with a real-time information retrieval component [4]. At inference time the system fetches relevant document passages from an external corpus and injects them into the prompt as a numbered retrieved-context block, giving the model a grounded context from which to synthesize its answer. RAG has been shown to reduce hallucination and enable citation of sources, making it a natural fit for high-stakes applications such as clinical decision support and biomedical literature synthesis [5, 6].

The retrieval component determines how much of that potential the system can actually realize. A retriever that returns irrelevant or redundant passages actively hurts generation quality, regardless of how capable the LLM is. The design space for retrievers has expanded considerably over the past few years: dense bi-encoder search [7, 8], sparse term-matching methods such as BM25 [9], hybrid combinations of both [10], cross-encoder reranking [11], LLM-based query expansion [12], and diversity-aware selection via Maximal Marginal Relevance (MMR) [13] have each attracted significant research attention. Despite this breadth of options, very few studies directly compare multiple strategies within a single, controlled biomedical RAG pipeline while simultaneously reporting retrieval-quality and generation-quality metrics at the level of statistical confidence intervals.

This paper fills that gap. We construct a modular RAG benchmark on the BioASQ dataset [14]—a well-established testbed for biomedical question answering—and evaluate five retrieval strategies head-to-head under identical generation and embedding conditions. Our evaluation framework is built on DeepEval [15], which provides four complementary metrics: contextual precision (signal-to-noise ratio in the retrieved context), contextual recall (coverage of necessary evidence), faithfulness (factual consistency of the generated answer with the provided context), and answer relevancy (responsiveness of the answer to the question). We also include a no-context ablation that quantifies the value of retrieval as a whole, and we report 95% confidence intervals for every mean to distinguish genuine strategy differences from sampling noise.

The specific contributions of this work are:

- An end-to-end, reproducible RAG evaluation pipeline—covering data loading, chunking, embedding, retrieval, generation, and metric computation—whose every hyperparameter is version-controlled in a single configuration file.
- A controlled empirical comparison of five retrieval strategies on 250 BioASQ QA pairs, with per-metric means and 95% confidence intervals, supported by a no-context lower bound.
- Actionable, evidence-grounded guidance on strategy selection for practitioners building biomedical RAG systems, comparing contextual precision, contextual recall, faithfulness, and answer relevancy across strategies.

## 2 Related Work

### 2.1 Retrieval-Augmented Generation

Lewis et al. [4] introduced RAG as a general framework for knowledge-intensive NLP tasks, pairing a dense retriever with a seq2seq generator. Subsequent surveys [10, 16] have catalogued the rapid diversification of the approach, from single-hop retrieval to iterative, self-reflective pipelines [17]. In the biomedical domain specifically, RAG has been applied to clinical QA [5], drug discovery literature synthesis [6], and semantic indexing. Our work complements this literature by isolating the retrieval component as the sole experimental variable within a fixed biomedical QA setting.

### 2.2 Information Retrieval Methods

Sparse retrieval via BM25, formalized by Robertson and Zaragoza [9], remains competitive because it matches exact medical terms—drug names, gene identifiers, anatomical structures—without depending on training-data coverage. Dense retrieval, popularized by DPR [7] and Contriever [8] and ANCE [18], captures semantic similarity through learned embeddings, handling paraphrases and synonymy that confound BM25. Hybrid approaches [10] attempt to capture both strengths; Reciprocal Rank Fusion (RRF) is a common score-combination technique that is parameter-light.

Cross-encoder reranking—first demonstrated at scale by Nogueira and colleagues [11, 19]—refines an initial candidate pool by jointly encoding query and document tokens, enabling token-level attention across both. This richer interaction has linear cost in the candidate pool size, making it suitable as a second stage that operates over a small pool rather than the full corpus. Multi-query expansion [12] addresses query ambiguity by generating several rephrased versions of the original query, retrieving independently for each, and returning a deduplicated pool ranked by how frequently each passage was retrieved across variants. Carbonell and Goldstein [13] proposed MMR to reduce redundancy in retrieved sets through iterative selection of diverse, relevant candidates.

### 2.3 Biomedical Question Answering

BioASQ [14] is a long-running benchmark for large-scale biomedical semantic indexing and question answering, drawing questions from domain experts and grounding answers in PubMed abstracts. The rag-mini-bioasq subset we use provides pre-curated corpus and QA pairs in a format designed for RAG pipeline evaluation. Complexity in biomedical language—polysemy, dense use of abbreviations, and fine-grained entity relationships—makes it a demanding testbed; Soong et al. [6] explicitly showed that retrieval quality has an outsized impact on LLM accuracy for biomedical queries relative to open-domain settings.

### 2.4 Automated RAG Evaluation

Human evaluation of RAG systems is expensive and difficult to replicate. RAGAS [20] and DeepEval [15] both provide LLM-based automatic metrics that decompose quality into complementary dimensions. Faithfulness, in both frameworks, measures whether a generated answer makes only claims supported by the provided context—a critical property for safety-sensitive applications. Answer relevancy captures whether the response actually addresses the question posed. These automatic metrics have known limitations [21], particularly when the same model acts as both generator and judge (a circularity we discuss in Section 7); nonetheless they provide reproducible, scalable signal that correlates with human judgment. Reimers and Gurevych [22]'s work on sentence embeddings underpins the similarity calculations used within several of these metrics.

## 3 System Design and Methodology

### 3.1 Pipeline Architecture

The evaluation pipeline proceeds through four sequential phases: (1) data ingestion and validation, (2) embedding and vector index construction, (3) per-strategy retrieval and generation, and (4) metric computation and visualization. Each phase is implemented as a standalone Python script that reads from a single YAML configuration file. This design—inspired by reproducibility requirements articulated in the machine learning community—means any researcher can re-run the exact experiment by changing a single file rather than hunting for hard-coded constants scattered across a codebase.

The retrieval strategy is the only component that varies across experimental conditions. Embedding model, vector store, generation model, prompt template, chunking parameters, and evaluation sample are held constant. LangChain v0.2.17 [23] provides the integration layer that makes this modular swap-out practical: each strategy is encapsulated behind a common *BaseRetriever* interface, so the downstream pipeline code is completely blind to which strategy is active.

### 3.2 Dataset

We use the *rag-mini-bioasq* dataset (HuggingFace: *enelpol/rag-mini-bioasq*) [14], a curated RAG-ready subset of the BioASQ challenge corpus. The dataset provides 707 question-answer pairs in the evaluation split, paired with a corpus of 40,181 PubMed-derived passages. Dataset integrity is verified at pipeline startup: all passages referenced by QA pairs must exist in the corpus, and 99.2% of QA pairs require synthesis of information from multiple passages (i.e., answers are not verbatim-extractable from any single passage), confirming that retrieval quality directly determines generation quality for virtually all questions in this dataset. We randomly sample 250 QA pairs (seed = 42) for evaluation; this sample size provides 95% confidence intervals of approximately ±0.04 width on proportions, sufficient to detect meaningful between-strategy differences. The rag-mini-bioasq dataset is distributed under the Creative Commons Attribution 2.5 (CC BY 2.5) license; our use of this dataset is in full compliance with its terms.

### 3.3 Text Chunking

The 40,181 corpus passages are split into chunks using LangChain’s *RecursiveCharacterTextSplitter* with a chunk size of 512 tokens (cl100k_base encoding) and an overlap of 50 tokens (~10%). The splitter uses a priority-ordered separator hierarchy: paragraph breaks, line breaks, sentence boundaries, then word boundaries, preserving semantically coherent units wherever possible. The resulting index contains 45,491 chunks—an average of roughly 1.132 chunks per source passage. The 10% overlap mitigates boundary artifacts where a relevant sentence straddles two consecutive chunks. This chunk size is consistent with findings from Wang et al. [24], who demonstrated a precision-recall optimum near 512 tokens for biomedical retrieval.

### 3.4 Embedding Model and Vector Store

All text representations are generated using OpenAI's *text-embedding-3-small* model [25], which produces 1,536-dimensional vectors. We chose this model for its strong performance on semantic similarity benchmarks and its practical cost advantage over larger embedding models—an important consideration given that the full corpus must be embedded once and QA-pair embeddings must be generated at evaluation time for each of the five strategies. Domain-specific alternatives such as BioBERT or SciBERT embeddings [26, 27] represent a natural extension but are out of scope for the present controlled comparison, whose goal is strategy isolation rather than embedding-model tuning.

Embeddings are stored in ChromaDB v0.4.24 [28], an open-source, embedded vector database that requires no separate server process. ChromaDB supports cosine-similarity k-NN search and integrates natively with LangChain. Index provenance metadata (embedding model name, chunk size) is verified at load time to prevent accidental cross-contamination between experimental configurations.

### 3.5 Retrieval Strategies

All strategies retrieve $k = 10$ passages per query. This value is held constant across all conditions. We describe each strategy's mechanism, then discuss its intended strengths and known failure modes in the biomedical context.

Dense Vector Search (Baseline). Cosine-similarity k-nearest-neighbor search over the ChromaDB index. The query is embedded with text-embedding-3-small and the 10 chunks with the highest cosine similarity are returned. This is the canonical RAG retriever and serves as our experimental baseline [7].

Hybrid BM25 + Dense. LangChain's EnsembleRetriever combines BM25 [9] (weight 0.4) and dense retrieval (weight 0.6) through Reciprocal Rank Fusion. BM25 operates on a full-corpus inverted index built from the same 40,181 passages. The rationale is that biomedical queries frequently contain exact clinical terms—drug names, gene identifiers, anatomical abbreviations—where lexical matching complements semantic search [10].

Cross-Encoder Reranking. A two-stage pipeline first retrieves 30 candidates via dense search (initial_k = 3k = 30), then re-scores every (query, passage) pair using the *cross-encoder/ms-marco-MiniLM-L-6-v2* model [11]. Unlike bi-encoders, the cross-encoder attends jointly over query and passage tokens, producing relevance scores that capture precise query-passage interaction. The top 10 passages by cross-encoder score are returned. This architecture is deliberately asymmetric: the expensive cross-encoder operates on a small candidate pool, while cheap dense search pre-filters the full corpus.

Multi-Query Expansion. GPT-4o-mini generates three alternative phrasings of the original query. Dense retrieval runs independently for each phrasing, and results are merged with deduplication. Passages are ranked by the number of query variants that retrieved them (cross-variant frequency), with document ID as a deterministic tie-breaker [23] (an approach conceptually related to RAG-Fusion [12]). The motivation is to address query ambiguity common in clinical language (e.g., "MI" for myocardial infarction vs. mitral insufficiency) by triangulating from multiple semantic angles.

Maximal Marginal Relevance (MMR). ChromaDB's native MMR implementation fetches fetch_k = 40 candidates (fetch_k_multiplier = 4) and then iteratively selects passages that balance relevance to the query against similarity to already-selected passages, controlled by a balance parameter $\lambda = 0.5$ (following LangChain's convention where $\lambda = 0$ maximises relevance and $\lambda = 1$ maximises diversity, which is the inverse of the original formulation in [13]). The goal is to reduce semantic redundancy common in biomedical literature, where multiple passages from the same publication often cover the same ground.

### 3.6 Generation

Answer generation uses GPT-4o-mini (gpt-4o-mini-2024-07-18) [29] (temperature 0.1, max_tokens 500, 3 retries). The low temperature reflects the domain requirement for factual consistency rather than creative elaboration. A fixed system prompt instructs the model to: (1) base its answer exclusively on the retrieved passages, (2) include inline citations referencing the numbered passages, (3) use precise biomedical terminology, and (4) append a safety disclaimer noting that responses are informational and not medical advice. The explicit grounding instruction—"If the documents don't contain enough information, say 'Insufficient evidence in provided sources'"—is critical for faithfulness; it prevents the model from supplementing retrieved context with parametric knowledge. The complete system prompt is reproduced verbatim below:

```
You are a biomedical research assistant. Your task is to answer questions based
ONLY on the provided document excerpts.

CRITICAL INSTRUCTIONS:

1. GROUNDING: Base your answer entirely on the retrieved documents below. If
the documents don't contain enough information, say "Insufficient evidence in
provided sources."
2. CITATIONS: Cite each claim using [1], [2], etc., referencing the numbered
documents. Example: "ACE inhibitors reduce blood pressure [1][3]."
3. CLARITY: Use precise biomedical terminology but explain for educated lay
audiences.
4. SAFETY: This is for informational purposes only. DO NOT provide medical
advice. Include disclaimer: "Not medical advice. Consult healthcare
professionals."
5. HALLUCINATION: Do not add information not present in the documents. Do not
speculate.

Retrieved Documents:
{context}

Question: {question}

Answer with citations and safety disclaimer
```

### 3.7 Evaluation Framework

We evaluate each (question, retrieved_context, generated_answer) triple using four DeepEval v1.4.9 [15] metrics:

- Contextual Precision: the fraction of retrieved chunks that are relevant to the query (signal-to-noise ratio in the context window).
- Contextual Recall: whether the retrieved context contains all information needed to answer the question (coverage of necessary evidence).
- Faithfulness: whether every claim in the generated answer is supported by the provided context (factual grounding).
- Answer Relevancy: how directly the generated answer addresses the original question (response utility).

All four metrics are computed by GPT-4o-mini (gpt-4o-mini-2024-07-18) acting as an LLM judge with temperature 0.0. The judge model is separate in configuration from the generation model—in practice both resolve to the same model identifier, a circularity we acknowledge as a limitation (Section 7). Scores are bounded in [0, 1]. We report per-strategy means and 95% bootstrap confidence intervals computed over the 250 evaluation samples. Strategy-level scores are also aggregated into a composite score (unweighted mean of the four metrics) to enable single-number ranking. A no-context condition—where the pipeline runs without any retrieved passages—serves as an ablation baseline.

## 4 Experimental Setup

All retrieval and evaluation experiments were executed under run ID 2026-03-29_21-17. The corpus vector index was pre-built in a prior indexing run and loaded from cache for all strategy evaluations. The global random seed is set to 42 across Python's *random* module, NumPy, and the evaluation sampler, ensuring the 250 QA pairs drawn are identical across all six strategy conditions. Table 1 summarizes the key hyperparameters.

| Parameter | Value |
|---|---|
| Embedding model | text-embedding-3-small (1536-dim) |
| Vector store | ChromaDB (cosine similarity) |
| Chunking | 512 tokens, 50-token overlap |
| Generator | gpt-4o-mini-2024-07-18 (T=0.1, max_tokens=500) |
| Judge model | gpt-4o-mini-2024-07-18 (T=0.0) |
| Top-k retrieved (k) | 10 |
| Reranker initial pool | 30 (cross-encoder strategy only) |
| BM25 weight | 0.4 (hybrid strategy only) |
| MMR lambda | 0.5 (MMR strategy only) |
| Multi-query variants | 3 (multi-query strategy only) |
| Evaluation sample size | 250 QA pairs per strategy |
| Random seed | 42 |

*Table 1. Experimental hyperparameters.*

Strategy evaluation proceeds sequentially: context retrieval is batched first across all 250 questions, then generation and metric computation run in batches of 50 with a 20-second inter-batch pause to respect API rate limits. Intermediate results are checkpointed every 50 QA pairs to allow resumption in case of API interruption. The judge model temperature is set to 0.0 to ensure deterministic scoring; because DeepEval handles judge model invocation internally, this temperature is requested through the framework's configuration rather than direct API calls.

All experiments were run locally on a MacBook Pro with an Apple M4 Max chip. Total OpenAI API cost across all five strategies and the no-context baseline was approximately $11.00 USD. End-to-end wall-clock time for the full experimental run—covering embedding construction, retrieval, generation, and evaluation—was approximately 5.5 hours, with the majority of elapsed time attributable to API latency during the DeepEval metric computation phase.

## 5 Results

Table 2 presents the complete evaluation results, Figure 1 visualizes the per-strategy, per-metric scores as a heatmap, and Figure 2 shows the composite score ranking. Every RAG strategy scores markedly higher than the no-context ablation on answer relevancy (0.658–0.701 vs. 0.287), confirming that retrieval provides a measurable benefit in this domain. We discuss each metric and notable strategy-level patterns in turn.

| Strategy | Ctx. Prec. | Ctx. Recall | Faithful. | Ans. Relev. | Composite |
|---|---|---|---|---|---|
| Dense Vector Search (k-NN) | 0.809 [0.766, 0.846] | 0.887 [0.856, 0.921] | 0.897 [0.875, 0.918] | 0.695 [0.661, 0.726] | **0.822** |
| Hybrid BM25 + Dense | 0.817 [0.782, 0.853] | 0.874 [0.839, 0.907] | 0.893 [0.870, 0.914] | 0.696 [0.664, 0.731] | **0.820** |
| Cross-Encoder Reranking | 0.852 [0.818, 0.885] | 0.852 [0.814, 0.889] | 0.902 [0.879, 0.922] | 0.701 [0.669, 0.737] | **0.827** |
| Multi-Query Expansion | 0.671 [0.630, 0.709] | 0.839 [0.804, 0.877] | 0.897 [0.876, 0.917] | 0.680 [0.647, 0.714] | **0.772** |
| Maximal Marginal Relevance | 0.824 [0.788, 0.859] | 0.775 [0.728, 0.813] | 0.893 [0.871, 0.913] | 0.658 [0.620, 0.692] | **0.787** |
| *No Context (Parametric)* | 0.000 [0.000, 0.000] | 0.000 [0.000, 0.000] | 0.978 [0.965, 0.990] | 0.287 [0.246, 0.329] | 0.316 |

*Table 2. Mean metric scores with 95% confidence intervals (n = 250). Ctx. Prec. = Contextual Precision; Ctx. Recall = Contextual Recall; Faithful. = Faithfulness; Ans. Relev. = Answer Relevancy. Composite is the unweighted mean of all four metrics (RAG strategies only). †No-context faithfulness is high by construction: DeepEval scores faithfulness against retrieved context; with no context provided, there is no material for the model to contradict.*

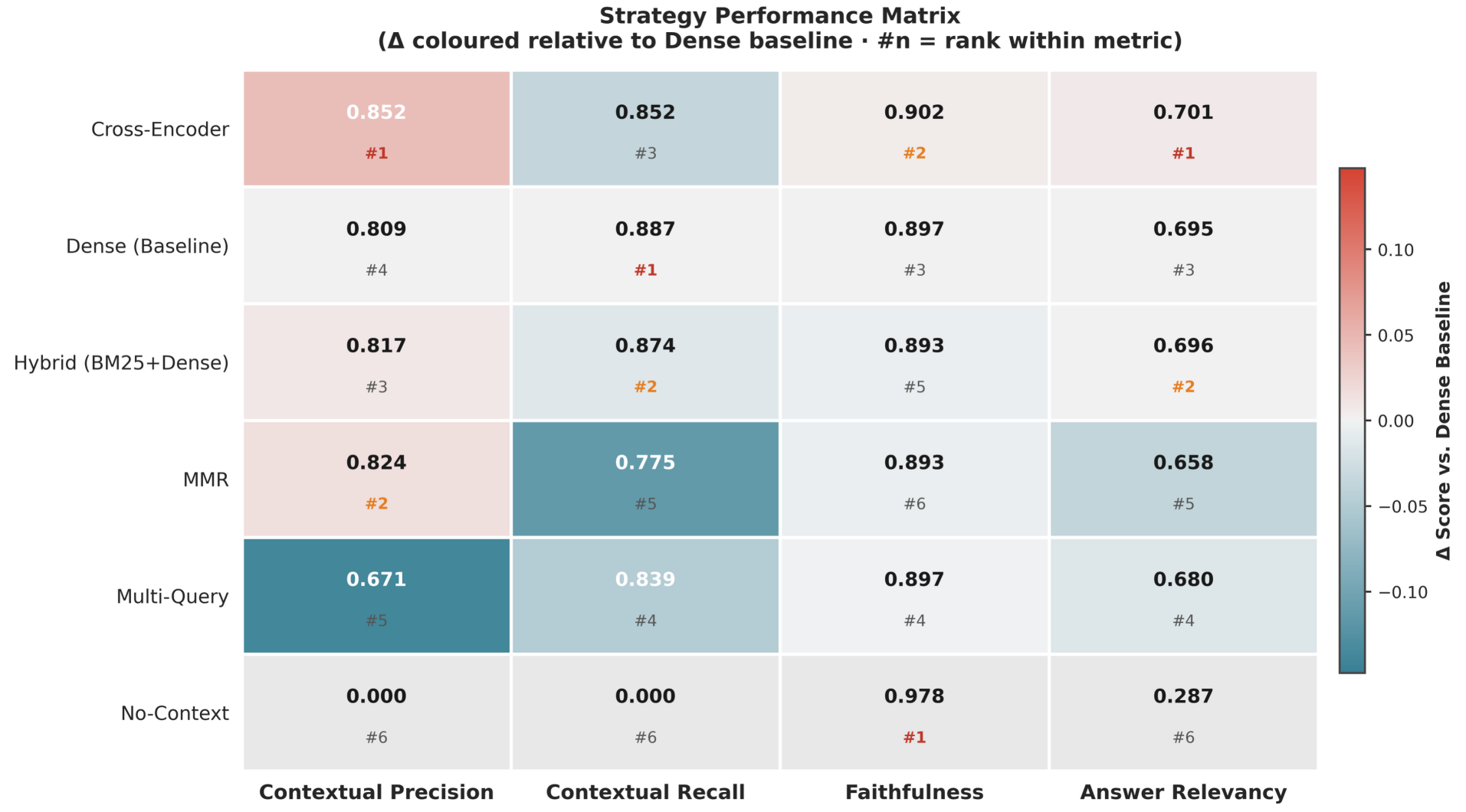

*Figure 1. Strategy Performance Matrix. Each cell shows the mean metric score and within-column rank (#1 = best across all six strategies); colour encodes the delta relative to the Dense baseline (red = above baseline, blue = below). Cross-Encoder ranks #1 in contextual precision (0.852) and answer relevancy (0.701). Dense (Baseline) ranks #1 in contextual recall (0.887). No-Context ranks #1 in faithfulness by construction (0.978, an artefact of the metric definition) but last on all retrieval-dependent metrics.*

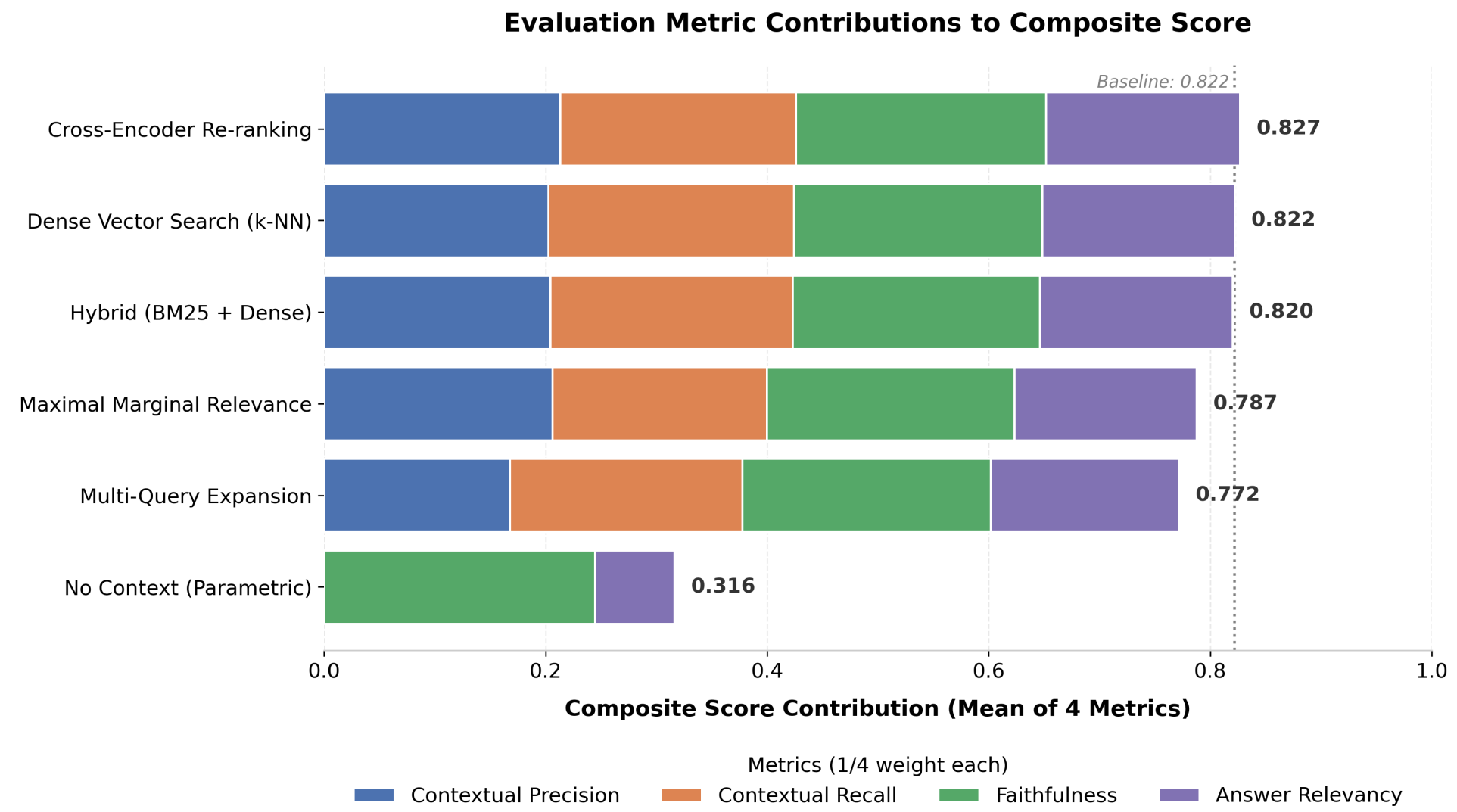

*Figure 2. Evaluation metric contributions to composite score (unweighted mean of four metrics, each contributing 1/4 weight). Cross-Encoder Re-ranking leads at 0.827, followed by Dense Vector Search (k-NN) at 0.822 (the dotted vertical line marks the Dense baseline) and Hybrid (BM25 + Dense) at 0.820. Maximal Marginal Relevance scores 0.787, Multi-Query Expansion 0.772, and No Context (Parametric) 0.316. Bar segments show the per-metric contribution of Contextual Precision (blue), Contextual Recall (orange), Faithfulness (green), and Answer Relevancy (purple).*

## 5.1 Contextual Precision

Cross-Encoder Reranking achieves the highest contextual precision at 0.852 (95% CI [0.818, 0.885]). Multi-Query Expansion is the clear outlier, scoring 0.671 [0.630, 0.709]—a difference of 18 percentage points from the top strategy and well outside the confidence intervals of every other strategy. The remaining three approaches cluster closely: MMR (0.824), Hybrid (0.817), and Dense (0.809). The narrow gap among these three, with overlapping confidence intervals, prevents any strong ordering claim beyond the Cross-Encoder's superiority.

The Multi-Query result runs against expectation: expanding the query was intended to broaden recall without hurting precision, yet the opposite occurs. We hypothesize that the three LLM-generated paraphrases often retrieve thematically adjacent but query-specific irrelevant passages, which then enter the merged context without a reranking step to filter them. The cross-variant frequency ranking used for deduplication (passages appearing in more query variant results rank higher) does not distinguish between truly relevant passages and passages that are generically popular across semantically broad paraphrases.

## 5.2 Contextual Recall

Dense Vector Search leads on recall (0.887 [0.856, 0.921]), with the Hybrid strategy close behind (0.874 [0.839, 0.907]). Cross-Encoder Reranking (0.852 [0.814, 0.889]) and Multi-Query (0.839 [0.804, 0.877]) form a middle tier. MMR shows the steepest recall penalty at 0.775 [0.728, 0.813]—a direct consequence of its diversity objective: by actively suppressing semantically similar passages, it occasionally excludes passages that contain complementary evidence for the same question.

The recall ordering reveals the precision-recall tradeoff inherent in retrieval design. Cross-Encoder Reranking improves precision by 4.3 points over Dense but costs 3.5 recall points, while MMR improves precision by 1.5 points over Dense at a cost of 11.2 recall points—a poor trade at the default lambda of 0.5. Figure 3 shows the per-metric delta from the Dense baseline, making these tradeoffs visually explicit.

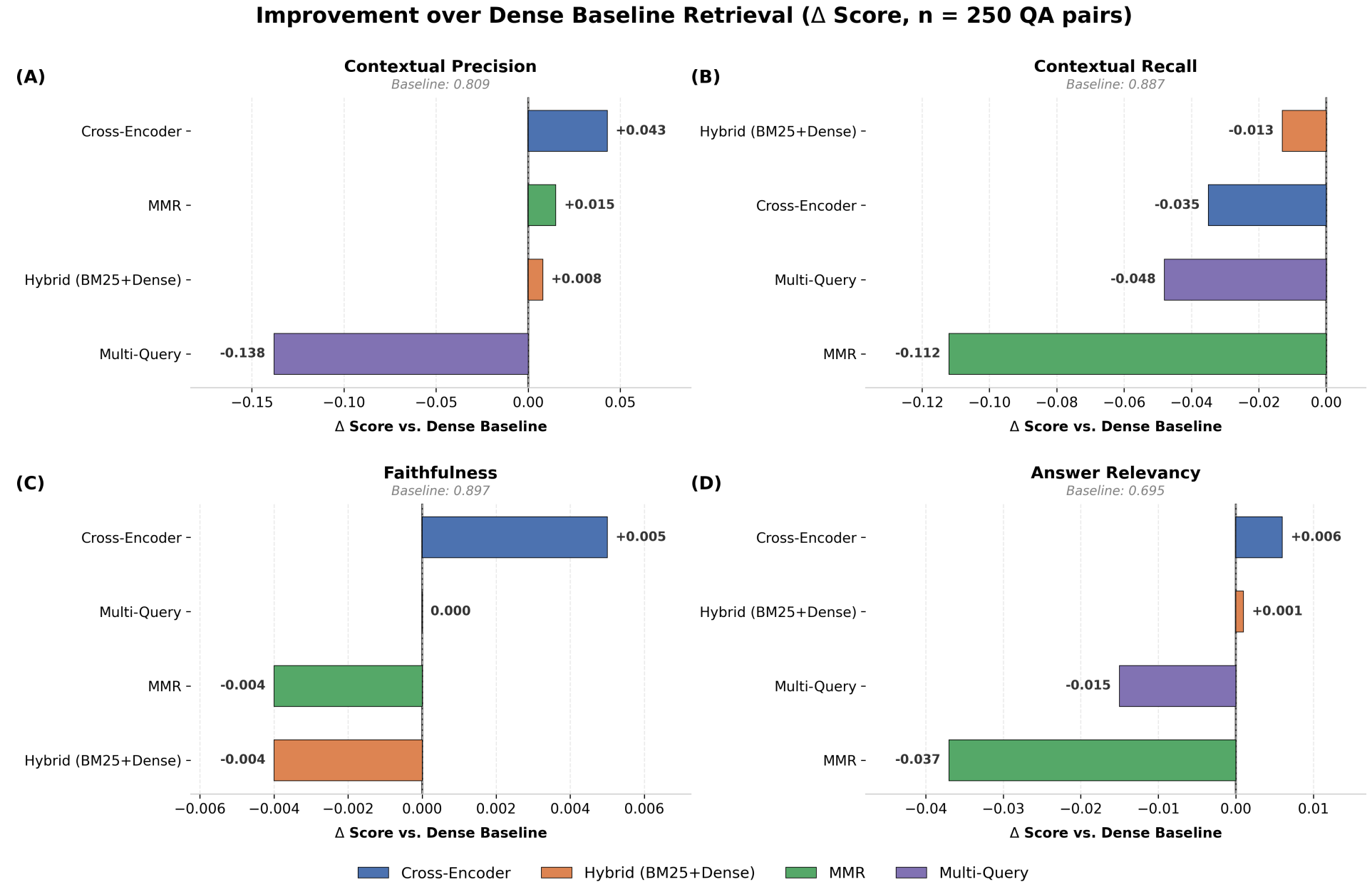

*Figure 3. Per-metric improvement over the Dense Vector Search baseline (n = 250 QA pairs). (A) Contextual precision (baseline: 0.809): Cross-Encoder gains +0.043; Multi-Query suffers the largest loss (−0.138). (B) Contextual recall (baseline: 0.887): all strategies decline, with MMR posting the steepest loss (−0.112). (C) Faithfulness (baseline: 0.897): differences are negligible across strategies (range −0.004 to +0.005). (D) Answer relevancy (baseline: 0.695): Cross-Encoder gains +0.006; MMR loses −0.037.*

### 5.3 Faithfulness

Faithfulness scores are uniformly high across all RAG strategies, ranging from 0.893 to 0.902, with no statistically distinguishable differences given overlapping confidence intervals. Cross-Encoder Reranking achieves the highest mean (0.902 [0.879, 0.922]) and Multi-Query is statistically indistinguishable from Dense (both 0.897). The no-context ablation records 0.978 on faithfulness, which is a measurement artifact rather than a genuine result: DeepEval evaluates faithfulness as the absence of claims that contradict the provided context; with no context, there is nothing to contradict, and the score defaults to near-perfect. This shows why faithfulness alone is an unreliable quality signal for RAG systems—it must be read alongside answer relevancy.

The uniformly high faithfulness values across RAG strategies suggest that the generation prompt—which explicitly instructs the model to use only retrieved passages and to flag insufficient evidence—is effective at suppressing hallucination regardless of which retrieval strategy supplies the context. This is an important practical finding: a well-designed grounding prompt can maintain faithfulness even when retrieved context quality varies.

### 5.4 Answer Relevancy

Answer relevancy shows the widest spread among the four metrics, ranging from 0.658 (MMR) to 0.701 (Cross-Encoder Reranking). All RAG strategies improve dramatically over the no-context condition (0.287), confirming that the primary driver of answer relevancy is retrieval quality rather than the model's parametric knowledge. The no-context system produces answers that are syntactically coherent but generically biomedical—they satisfy surface-level plausibility without actually addressing the specific question posed.

The MMR strategy’s lower answer relevancy (0.658 [0.620, 0.692]) reflects its diversity-first design: by pulling in passages that cover different aspects of a topic, it sometimes introduces context about adjacent concepts that shifts the generated answer away from the specific question. The radar chart in Figure 4 makes this asymmetry visually apparent. Whether this diversity is a feature or a bug depends on the use case—for exploratory biomedical queries it may be desirable, while for precision-critical clinical questions it is a liability.

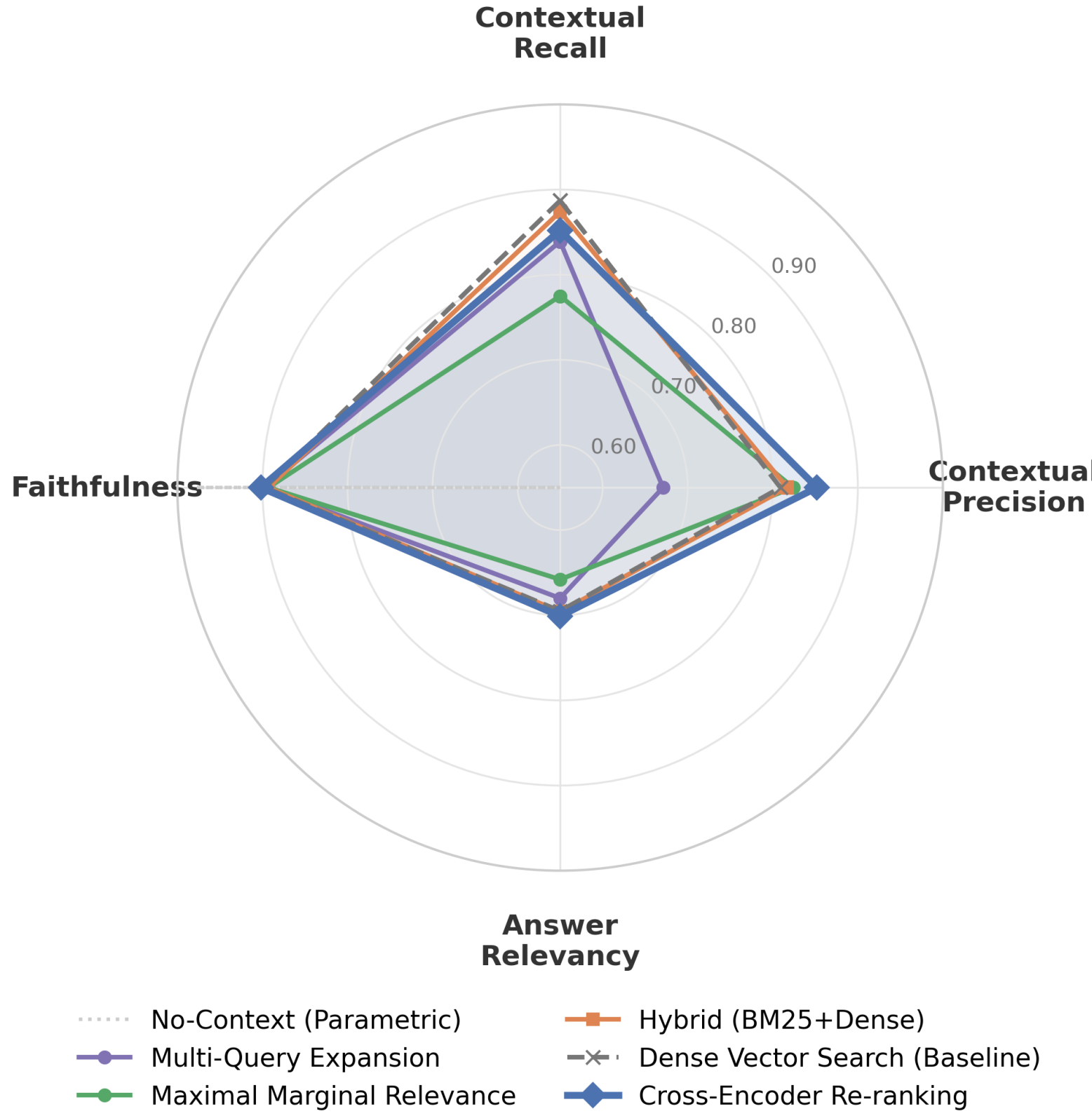

*Figure 4. Multi-metric performance profile (radar chart) comparing all six strategies across the four evaluation dimensions. Cross-Encoder Re-ranking occupies the largest total area, driven by its contextual precision advantage (0.852). Multi-Query Expansion shows the most asymmetric profile with the weakest contextual precision (0.671) of any retrieval strategy. MMR records the lowest contextual recall (0.775) and answer relevancy (0.658) among retrieval strategies. No-Context (Parametric) collapses to near-zero on contextual precision and recall.*

# 6 Discussion

## 6.1 Why Does Cross-Encoder Reranking Outperform?

The consistent edge of Cross-Encoder Reranking on precision—and its best-overall composite score (Figure 2)—can be attributed to the architectural difference between bi-encoders and cross-encoders. A bi-encoder compresses the full semantic content of a document into a single fixed-length vector before the query is seen; this compression necessarily loses query-specific relevance signals. A cross-encoder, by contrast, processes the query and passage together, allowing full token-level attention across both. In the biomedical domain, where a single term (e.g., "resistance") can refer to antibiotic resistance, drug resistance, electrical resistance, or psychological resistance, this query-conditioned reading matters. The cross-encoder can identify that a passage about "fluoroquinolone resistance mechanisms in E. coli" is highly relevant to the query "how do bacteria evade ciprofloxacin?" even if the overall cosine distance between their embeddings is moderate.

The two-stage design—30-candidate dense retrieval followed by cross-encoder rescoring—also means the reranker never sees irrelevant passages that did not survive the initial dense filter. This pre-filtering keeps cost manageable: the cross-encoder scores only k=30 candidates per query rather than the full corpus, making the computational overhead proportional to k, not to corpus size.

## 6.2 The Multi-Query Precision Paradox

Multi-Query Expansion's poor contextual precision (0.671) appears at first to contradict its design intent. The strategy is meant to improve recall by covering different semantic angles of a question. In practice, recall actually decreases modestly (0.839 vs. 0.887 for Dense), and precision deteriorates sharply. The most likely explanation is the corpus scale: with 45,491 chunks and queries that are already fairly specific biomedical questions, there is limited ambiguity to resolve. The LLM-generated paraphrases explore variations that the dense retriever would have captured anyway, and the additional retrieval sweeps primarily add noise—passages that are topically adjacent but do not answer the specific question. At larger corpus scales, or for shorter, more ambiguous queries, Multi-Query Expansion may show greater value.

## 6.3 Hybrid Search and the Limits of Naive Fusion

Hybrid BM25 + Dense retrieval was expected to outperform pure dense search by combining lexical exactness with semantic coverage. Instead, it scores just below Dense on both recall (0.874 vs. 0.887) and composite (0.820 vs. 0.822). The BM25 weight of 0.4 and dense weight of 0.6 were fixed without dataset-specific tuning; the optimal split likely depends on the lexical specificity of the queries in the evaluation sample. The rag-mini-bioasq questions are predominantly complex synthesis queries rather than exact-match lookups, which may explain why the BM25 component provides less lift than expected. Weight optimization or query-adaptive fusion could change this outcome noticeably.

## 6.4 The No-Context Ablation as a Sanity Check

The no-context condition functions as a lower bound and sanity check. Its near-zero answer relevancy (0.287) shows that GPT-4o-mini's parametric biomedical knowledge, while non-trivial, does not reliably produce on-target answers for BioASQ-level questions. The model's high faithfulness (0.978) in this condition is a cautionary note for evaluation design: faithfulness without an anchor context is meaningless, and any evaluation framework that reports faithfulness without contextual precision and recall risks creating a misleading picture of system quality.

## 6.5 Practical Guidance for System Builders

The five strategies studied here occupy distinct positions in the precision-recall space, and no single strategy dominates across all four metrics. Practitioners should anchor their choice on the constraint that is hardest to relax in their deployment context.

**When the primary constraint is clear:**

- *Precision-critical applications* (clinical decision support, evidence-based Q&A where irrelevant context actively harms generation): Cross-Encoder Reranking is the recommended choice. It achieves the highest contextual precision (0.852) and the best composite score of any strategy. The two-stage architecture—dense retrieval of k=30 candidates followed by cross-encoder rescoring—is expected to add per-query latency relative to single-pass approaches, though latency was not formally measured in this study.
- *Recall-critical applications* (systematic literature review, hypothesis generation where missing relevant passages is the primary risk): Dense Vector Search is the best-performing and simplest option, achieving the highest recall (0.887) at the lowest architectural complexity—a single bi-encoder pass with no reranking or query expansion stage.

**When balancing multiple constraints:**

The precision-recall trade-off between the top two strategies is modest: Cross-Encoder gains 4.3 percentage points in precision (0.852 vs. 0.809) at a cost of 3.5 percentage points in recall (0.852 vs. 0.887). In most biomedical production settings this trade-off favors Cross-Encoder unless retrieval throughput is a hard constraint. MMR provides a measurably different retrieval profile—prioritizing result diversity over redundancy—but at the cost of answer relevancy (0.658, lowest of the retrieval strategies). It suits exploratory workflows such as broad literature scoping or hypothesis generation, where topical coverage matters more than answering a specific question precisely.

**Strategies requiring caution at this scale:**

- *Multi-Query Expansion* should not be the default choice for domain-specific corpora of comparable scale. It offered no recall advantage over Dense (0.839 vs. 0.887) while measurably degrading precision (0.671), and it triples the number of LLM API calls per query, increasing cost proportionally. It may prove more valuable in open-domain settings with larger, more heterogeneous corpora or with shorter, more ambiguous queries.
- *Hybrid BM25 + Dense* underperformed pure Dense retrieval under fixed weights (recall 0.874 vs. 0.887, composite 0.820 vs. 0.822). Naive fixed-weight fusion does not reliably outperform the dense baseline on synthesis-heavy biomedical queries. Dataset-specific weight tuning—or query-adaptive fusion that adjusts the BM25/dense ratio based on query lexical specificity—would be needed before adopting this strategy in a production setting.

## 7 Limitations

Several limitations bound the scope of our conclusions.

**Judge-generator circularity.** The same model (GPT-4o-mini) serves as both the answer generator and the DeepEval judge. This creates a potential circularity: the judge may systematically favor the writing style or reasoning patterns of its own outputs. Cross-judge evaluation—using a different model (e.g., GPT-4o or Claude Sonnet) as the judge—would provide more independent quality estimates and is a natural direction for follow-on work.

**Dataset scope.** Evaluation is conducted on a single curated subset of the BioASQ benchmark. The rag-mini-bioasq corpus is relatively modest in scale (40,181 passages). Strategy rankings may shift at larger corpus scales, with different question distributions, or on other biomedical tasks such as clinical note summarization or rare disease QA.

**Fixed embedding model.** All strategies share text-embedding-3-small embeddings. Domain-specific models such as BioBERT [26] may change the relative ordering of strategies, particularly for Hybrid search, where BM25's lexical component interacts with the semantic granularity of the dense retriever.

**Lack of human evaluation.** Automatic LLM-based metrics are proxies for human judgment. Their correlation with expert biomedical assessors has not been validated in this setting. A human evaluation study—particularly for faithfulness and answer relevancy in clinical contexts—would be necessary before deploying these rankings in a production clinical system.

**Strategy hyperparameter sensitivity.** Each strategy was evaluated at a single hyperparameter configuration. BM25 weight, MMR lambda, the number of Multi-Query variants, and the cross-encoder candidate pool size were set by design principle rather than dataset-specific optimization. A systematic grid search over these parameters would establish tighter performance bounds and could improve underperforming strategies noticeably.

**Latency not reported.** The experiment did not collect controlled wall-clock latency measurements; all timing is confounded by API rate-limit pauses, network jitter, and batching overhead. Future work should measure per-query latency under controlled conditions to enable a full performance-cost Pareto analysis.

**Chunking strategy.** All strategies operate on the same fixed chunking configuration (512-token chunks with 50-token overlap). Finer or coarser granularity may alter absolute metric values and could interact differently with each retrieval strategy—particularly Hybrid search, where BM25 term-frequency statistics are sensitive to chunk length.

## 8 Conclusion

We present a controlled empirical comparison of five retrieval strategies—Dense Vector Search, Hybrid BM25 + Dense, Cross-Encoder Reranking, Multi-Query Expansion, and Maximal Marginal Relevance—within a fully reproducible biomedical RAG pipeline evaluated on the BioASQ benchmark. The central finding is that increased retrieval complexity does not reliably translate into superior performance at this corpus scale. Cross-Encoder Reranking achieves the highest overall quality (composite 0.827) and the highest contextual precision (0.852), confirming that query-conditioned passage scoring yields meaningful gains in a domain where a single term can carry multiple clinical meanings. Yet Dense Vector Search is a formidable baseline (composite 0.822), falling only 0.005 points behind on the composite metric while achieving the highest recall (0.887) of any strategy. Hybrid BM25 + Dense and Multi-Query Expansion—both designed to outperform pure dense retrieval—finish below it: Multi-Query posts the lowest precision of any retrieval strategy (0.671), illustrating how query diversification without a subsequent reranking stage primarily introduces noise on specific, domain-constrained queries.

The no-context ablation provides a clear causal anchor: removing retrieval causes answer relevancy to collapse from approximately 0.66–0.70 to 0.287, even with GPT-4o-mini as the generation model, confirming that retrieval—not parametric knowledge—is the primary mechanism delivering answer quality in this setting. Uniformly high faithfulness scores across all RAG conditions (0.893–0.902) suggest that a grounding-focused system prompt reliably suppresses hallucination regardless of the underlying retrieval strategy, an encouraging signal for the safety requirements of clinical deployment.

Several directions merit follow-on investigation. Replacing text-embedding-3-small with domain-adapted biomedical encoders could alter the relative ordering of strategies, particularly for Hybrid and Cross-Encoder approaches. Dataset-specific tuning of the BM25 weight, MMR diversity parameter, and cross-encoder candidate pool size may narrow performance gaps that fixed-configuration evaluation cannot resolve. Cross-judge evaluation—using a model distinct from the answer generator—would establish whether the precision advantage of Cross-Encoder Reranking persists under independent assessment. Varying the chunking configuration (chunk size and overlap) would clarify how ingestion granularity interacts with each retrieval strategy. Finally, extending this benchmark to larger corpora and additional biomedical task types would test the generalizability of the strategy rankings reported here.

To support reproducibility and community extension, the complete pipeline code, all hyperparameter configurations, raw evaluation outputs, and visualization scripts are publicly available at https://github.com/deviprasadbal/RAGHealthcareRetrievalStrategies.

## Acknowledgements

Portions of the software implementation were developed with the assistance of AI-based programming tools. Experimental design, research methodology, results interpretation, and all scientific conclusions are solely the work of the authors. The BioASQ subset used in this study is available under its original license via HuggingFace. Computing costs were covered by the authors' OpenAI API credits.